\documentclass[accepted,specialissue]{melba}

\usepackage{mwe} 

\usepackage{amsmath,amsfonts,float,xurl}

\melbaid{2026:024}  
\doi{https://doi.org/10.59275/j.melba.2026-gc51}
\melbaauthors{Kats et al.}  
\email{eytan.kats@uni-luebeck.de}
\volume{2026}
\firstpageno{507}  
\melbayear{2026}  
\datesubmitted{2025-07-15}  
\datepublished{2026-07-22}  

\melbaspecialissue{MELBA–BVM 2025 Special Issue}
\melbaspecialissueeditors{Andreas Maier, Thomas Deserno, Heinz Handels, Klaus Maier-Hein, Christoph Palm, Thomas Tolxdorff, Katharina Breininger}

\ShortHeadings{Depth to Anatomy}{Kats et al.}

\title{Depth to Anatomy: Organ Localization from Depth Images for Automated Patient Table Positioning in Radiology Workflow}

\author{
	\firstname Eytan \surname Kats\aff{1}\orcid{0009-0004-3878-3036},
	\firstname Kai \surname Geissler\aff{2}\orcid{0009-0008-9709-0335},
    \firstname Jochen~G. \surname Hirsch\aff{2},
    \firstname Daniel \surname Mensing\aff{2}\orcid{0009-0009-8107-1313},
    \firstname Julien \surname Senegas\aff{3},
    \firstname Stefan \surname Heldman\aff{2},
    \firstname Mattias~P. \surname Heinrich\aff{1}\orcid{0000-0002-7489-1972}
}
\affiliations{
	\num 1 \addr Institute of Medical Informatics, University of Luebeck, Luebeck, Germany \\
	\num 2 \addr Fraunhofer Institute for Digital Medicine MEVIS, Bremen, Germany \\
    \num 3 \addr Philips Research, Hamburg, Germany
}

\abstract{
In clinical radiology, accurate patient table positioning is essential to align specific internal organs of interest with the scanner’s imaging isocenter, ensuring image quality and diagnostic reliability. Automated patient positioning can streamline this process and improve radiology workflow efficiency by reducing the time required for manual table adjustments and scout-based scan planning. We propose a learning-based framework that predicts 3D organ locations and shapes for 41 anatomical structures, including both bones and soft tissues, directly from a single 2D depth image of the body surface. Leveraging $10,020$ whole-body MRI scans from the German National Cohort (NAKO) dataset, we synthetically generate depth images paired with anatomical segmentations to train a convolutional neural network for volumetric organ prediction. Our method achieves a mean dice similarity coefficient of $0.44\pm0.2$ and and a symmetric average surface distance of $7.69\pm5.68$ mm across all structures. Furthermore, the model derives organ bounding boxes with a mean absolute detection offset of $10.99\pm5.54$ mm. Qualitative results on real-world depth images indicate the ability of the model to generalize to practical clinical settings. These findings suggest that depth-only organ localization can support automated patient positioning reducing setup time, minimizing operator variability, and improving patient comfort. The implementation and pretrained models are publicly available at \url{https://github.com/EytanKats/orgloc}.}


\begin{document}

\twocolumn[\maketitle]

\section{Introduction}

\enluminure{T}{he} growing demand for radiological examinations highlights the need to optimize scanning workflows in order to reduce patient waiting times and improve operational efficiency. An important yet time-consuming component of this workflow is the scan preparation and planning phase, which plays a crucial role in ensuring image quality and diagnostic reliability \citep{van2020cinderellas}. According to \citep{streit2021analysis}, scan preparation requires on average 5.7 minutes, corresponding to 11\% of the total examination time. This phase typically involves manual patient positioning followed by scout imaging to enable geometric planning of the subsequent diagnostic acquisition. In routine clinical practice, a radiologic technologist first aligns the patient on the scanner table, often relying on external anatomical landmarks and laser positioning guides to approximate the region of interest. After positioning inside the scanner bore, low-resolution multi-planar scout images are acquired. These images are then used to verify patient positioning and to manually define the geometry of the diagnostic sequences, including slice orientation, field of view, and scan range \citep{koken2009towards}.

Despite recent advances in radiology hardware and software, the automation of patient positioning remains a largely underexplored area compared to other aspects of scan acceleration or reconstruction \citep{shafieizargar2023systematic, heckel2024deep}. Traditional workflows still rely heavily on the experience and judgment of radiologic technologists. This manual planning process is both operator-dependent and time-consuming, particularly in complex examinations involving multiple regions or when precise anatomical targeting is required. Furthermore, variability in technologist expertise and patient anatomy can lead to inconsistent positioning, necessitating repeat localizers or even additional diagnostic scans, thereby increasing overall examination time and reducing scanner throughput \citep{danilouchkine2005operator}.

Automated patient positioning systems, capable of accurately identifying and localizing the anatomical region of interest, hold considerable promise for streamlining radiology workflows by automatically adjusting the patient table to the suitable imaging position. By reducing reliance on manual table adjustments and anatomical landmark estimation, these systems can minimize inter-operator variability, decrease setup time, and mitigate the risk of positioning errors, improving both imaging efficiency and diagnostic consistency across radiology departments.

\subsection{Related Works}

\paragraph{Surface-based patient positioning systems.}
Among the emerging technological approaches for automating patient table positioning, RGB-D camera-based systems have demonstrated significant potential for enhancing clinical workflows \citep{booij2021automated, booij2019accuracy}. These systems utilize depth sensing to capture the external body surface, identifying anatomical landmarks that serve as proxies for internal regions of interest. This information enables the localization of key structures, which can be used to infer the appropriate imaging volume and adjust the patient table automatically prior to scan acquisition, potentially eliminating the need for scout imaging. For instance, \cite{incetan2020rgb} demonstrated the utility of RGB-D sensors in localizing regions such as the head and thorax with high spatial accuracy. Similarly, \cite{karanam2020towards} employed depth-aware landmark detection to support robust, contactless positioning in MRI and CT workflows. However, these approaches primarily focus on external positioning and do not explicitly infer the 3D locations or shapes of internal anatomy.

\paragraph{Parametric body-model-based internal anatomy estimation.}
Several approaches address the prediction of internal anatomy by first deriving an intermediate 3D representation of the body, which is then used to infer internal structures. Building on the integration of statistical human body models with internal anatomy, recent research has extended the widely adopted Skinned Multi-Person Linear (SMPL) model \citep{loper2015smpl} to incorporate internal organ representations. The original SMPL is a parametric, skinned 3D body model that encodes human pose and shape with a low-dimensional set of learned parameters, enabling the synthesis of realistic body meshes through linear blend skinning. Guo et al. \citep{guo2022smpl} proposed SMPL-A, which augments SMPL with deformable internal organ geometries, allowing organ shapes to adapt dynamically with body pose. While effective in modeling pose-aware internal anatomy, SMPL-A requires a patient-specific full-body mesh to initialize internal organ shapes, limiting its generalizability and applicability in clinical environments. Similarly, OSSO \citep{keller2022osso} and BOSS \citep{shetty2023boss} infer internal organ shapes and locations by optimizing anatomical plausibility within the SMPL framework based on external body surface meshes. However, these methods are constrained by a limited scope: OSSO focuses exclusively on the skeletal system, and BOSS models only 12 specific anatomical structures. Extending this paradigm, the HIT (Human Internal implicit Tissues) framework \citep{keller2024hit} employs a deep implicit function to predict continuous volumetric tissue distributions conditioned on a parametric body surface mesh. While HIT achieves anatomically coherent internal structures, it focuses on general tissue distributions rather than the localization of discrete, individual organs.

\paragraph{Estimation of internal anatomy from depth images.}
Recent studies by \cite{geissler2025predicting} and \cite{kats2025internal} demonstrate that convolutional neural networks operating directly on depth maps can infer three-dimensional organ locations and approximate shapes from external body surface information. These works highlight that depth-derived surface cues contain sufficient anatomical information to estimate internal structures without explicit geometric reconstruction. While these works demonstrate that external body surface information contains useful cues about internal anatomy, their predictions are limited to planar projections rather than full volumetric organ localization. Similarly, \cite{teixeira2023automated} proposed a CNN-based method that estimates specific scan-planning parameters, such as lung centroids and patient isocenter, directly from depth images. Although effective for targeted task, such predictions provide only sparse anatomical descriptors and do not capture full organ extent or shape.

\begin{figure*}[t]
    \centering
    \includegraphics[width=0.76\linewidth, height=.36\textheight]{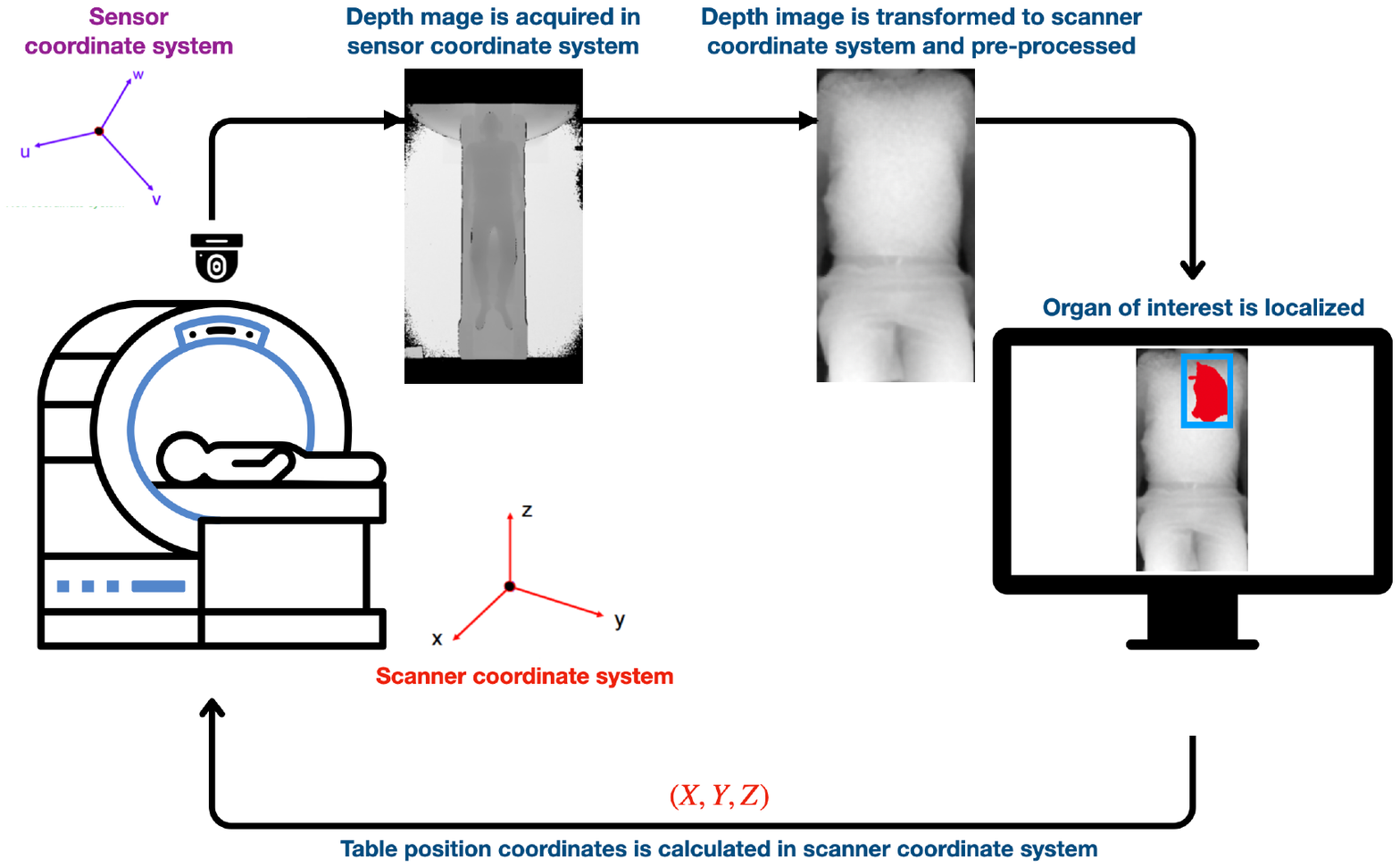}
    \caption{Overview of the proposed automated patient positioning workflow. A depth sensor mounted above the bore captures the patient's surface geometry. Through a spatial calibration and coordinate transformation process, the depth data is mapped into the scanner's coordinate system. The trained model predicts 3D organ segmentation within the scanner coordinate system. From this, a bounding box is derived to enable automated table translation, aligning the target anatomy with the imaging isocenter.}
    \label{fig_mri_workflow}
\end{figure*}

A closely related concurrent study is the LOOC framework \citep{henrich2025looc}, which infers internal organ locations and approximate shapes using an implicit volumetric occupancy model conditioned on depth-derived surface point clouds. The approach first maps the depth data to an external surface point cloud representation and subsequently predicts internal anatomy through a learned continuous occupancy field, enabling anatomically plausible volumetric estimates.

\subsection{Contribution}

This work presents a learning-based framework for estimating internal volumetric anatomical structures directly from a single anterior depth image. In contrast to surface-based patient positioning systems, which rely on external landmark detection or approximate region-of-interest estimation from body surface data \citep{booij2021automated, booij2019accuracy, incetan2020rgb, karanam2020towards}, our method explicitly predicts the three-dimensional locations and approximate shapes of internal anatomical structures. Unlike parametric body-model approaches that infer internal anatomy through template fitting or full-body surface mesh reconstruction within statistical body models \citep{guo2022smpl, keller2022osso, shetty2023boss, keller2024hit}, the proposed framework performs direct end-to-end depth-to-volume inference without intermediate body model fitting. Furthermore, whereas several recent depth-based approaches focus on predicting sparse anatomical descriptors such as organ centroids, isocenters, or limited projection-based organ localization \citep{teixeira2023automated, geissler2025predicting, kats2025internal}, our method produces explicit volumetric organ segmentations suitable for anatomically informed scan planning. Compared to the closely related concurrent LOOC framework \citep{henrich2025looc}, which employs an implicit volumetric occupancy model conditioned on reconstructed surface point clouds, our approach differs in methodology by directly predicting volumetric anatomy from depth images using a hybrid 2D-to-3D convolutional neural network. To support reproducibility and further research, the code and pretrained models are publicly available at~\url{https://github.com/EytanKats/orgloc}.

The key contributions of this work are summarized as follows:

\begin{enumerate}
    
    \item We introduce a hybrid 2D-to-3D convolutional neural network architecture (Pix2Vox) that maps a single anterior depth image to volumetric segmentations of 41 anatomical structures, providing explicit spatial information for automated radiological positioning.
    
    \item We demonstrate large-scale training and validation on a dataset comprising 10,020 MRI-derived paired depth and volumetric anatomy samples, enabling robust modeling of anatomical variability across a diverse population.
    
    \item We qualitatively show that a model trained purely on synthetic depth–anatomy pairs generalizes to real-world depth images of volunteers positioned in an MRI scanner, including scenarios with clothing, supporting the feasibility of practical clinical deployment.
    
\end{enumerate}

\section{Materials and Methods}
\label{sec_method}

In this study, we leverage $10,020$ whole-body MRI scans from the German National Cohort (NAKO) dataset \citep{bamberg2022whole}. This diverse dataset captures a broad spectrum of anatomical variability across age, sex, and body habitus, providing a foundation for training generalizable models of internal anatomy. As paired depth images and expert organ segmentations are not natively provided, we designed a pipeline to synthetically generate supervised training pairs directly from the MRI volumes.

To synthesize the input data, we simulate depth sensor acquisitions by rendering the anterior contours of the patient's body from the MRI volumes. This process utilizes the orthographic top-down projection to approximate a depth sensor mounted above the scanner table (Section~\ref{data_preparation}). To extract ground-truth 3D masks for 41 internal anatomical structures we utilize the TotalSegmentatorMRI tool \citep{d2024totalsegmentator} as detailed in Section~\ref{labels_preparation}.

The resulting dataset, comprising 2D depth images paired with 3D anatomical segmentations, is used to train a Pix2Vox convolutional neural network (Section~\ref{pix2vox}). This architecture is optimized to map 2D surface depth information to 3D volumetric predictions of the underlying target organs. Once trained, the model enables a depth sensor-augmented radiology workflow, facilitating automated patient positioning through anatomical structures localization (Section~\ref{rad_workflow}).

\subsection{Radiology Workflow with Integrated Depth Sensor}
\label{rad_workflow}

\begin{figure}[t]
    \centering
    \includegraphics[height=7cm]{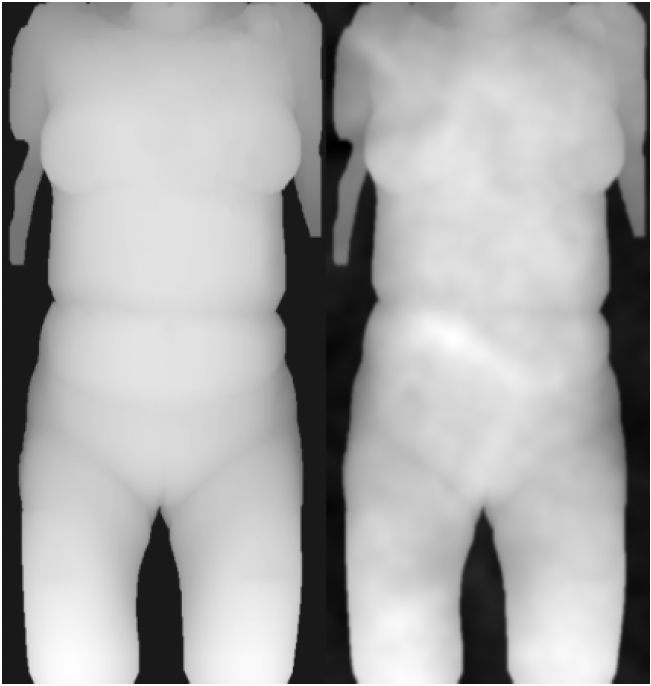}
    \caption{Example of creases and bulges simulation. Left: synthetic depth map extracted from an MRI scan. Right: augmented depth map featuring simulated creases and bulges.}
    \label{fig_augmentations}
\end{figure}

In the proposed workflow, a depth sensor is mounted above the scanner bore to capture an anterior depth image of the patient on the table. The sensor is spatially calibrated with the MRI system, allowing the acquired depth data to be transformed into a surface point cloud within the scanner’s coordinate frame \citep{booij2021automated, booij2019accuracy}. This point cloud is subsequently rendered as an orthographic depth projection (Section~\ref{real_depth_preparation}). The resulting processed image is then fed into the trained model, which predicts the 3D segmentations of the anatomical structures of interest in scanner space. Based on these predictions, the scanner table can be automatically translated to ensure the target anatomy is centered within the imaging isocenter, thereby minimizing the reliance on manual adjustments or iterative scout imaging (Figure~\ref{fig_mri_workflow}). This approach offers a contactless, fast, and standardized method of patient localization, which could be particularly beneficial in high-throughput clinical environments.

\begin{figure*}[t]
    \centering
    \includegraphics[width=0.98\linewidth]{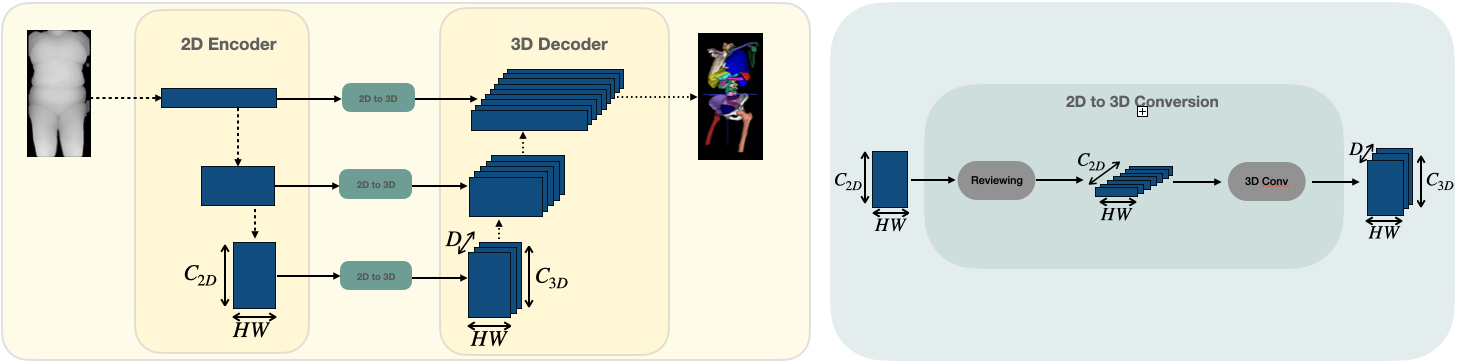}
    \caption{Overview of the proposed hybrid Pix2Vox network architecture. Left: The full model consists of a 2D encoder, 3D decoder, and custom 2D-to-3D conversion layers that connect the encoder and decoder at the bottleneck and each skip connection. Right: A detailed view of the 2D-to-3D conversion block. This module reshapes the 2D feature maps by unsqueezing a feature dimension and applying a 3D convolution to produce a volumetric representation suitable for further 3D processing. For clarity, the batch dimension is omitted from all illustrated tensor shapes.}
    \label{fig_model_arch}
\end{figure*}

\subsection{Synthetic Depth Image Generation}
\label{data_preparation}

To facilitate the training of a body surface-to-internal organs model, we designed a deterministic pipeline to synthesize 2D orthographic depth projections from 3D MRI volumes. The process transforms volumetric data $\mathcal{V} \in \mathbb{R}^{H \times W \times D}$ into depth maps $D(x, z)$

First, MRI intensity values are normalized to the range $[0, 1]$. A binary mask $M$ is generated by applying a low-intensity threshold $\epsilon = 0.02$ to reduce the inherent noise:
\begin{equation}
    M_{x,y,z} = [ \mathcal{V}_{x,y,z} > 0.02 ]
\end{equation}
To remove floating artifacts and ensure structural continuity of the skin surface, we apply binary morphological opening $\gamma_{B}(M)$ using a spherical structuring element with diameter equal to 3. 

The depth image $D$ is generated via a ray-casting approach along the anterior-posterior axis. For each lateral and longitudinal coordinate $(x, z)$, the distance to the first encountered foreground voxel is computed:
\begin{equation}
    d(x, z) = \text{argmax}_{y} (M_{x,y,z})
\end{equation}
The raw indices are normalized by the maximum volume height $H_{max}$ and inverted so that the resulting depth image pixel intensity values are inversely proportional to the physical distance from the sensor:
\begin{equation}
    D_{norm}(x, z) = 1 - \frac{d(x, z)}{H_{max}} \quad \text{for } d(x, z) > 0
\end{equation}

To clean signal from the scanner table and hardware artifacts, we apply a clipping threshold, zeroing all pixels with $D_{norm} < 0.3$. High-frequency surface noise and acquisition spikes are suppressed using grayscale morphological opening with an $11 \times 11$ kernel.

To bridge the domain gap between clean synthetic depth maps and real-world clinical acquisitions, we implemented a stochastic augmentation pipeline designed to simulate the depth profile of hospital gowns, blankets, and clothing. The augmented depth map $D_{aug}$ is generated by modifying the synthetic depth $D$ within the patient silhouette mask $S$ as follows:

\begin{equation}
    D_{aug} = G_{\sigma} \left( D + [B_{creases} + B_{bulges}] \cdot S \right)
\end{equation}

where $G_{\sigma}$ denotes a global Gaussian smoothing kernel with $\sigma=5$, $B_{creases}$ simulates creases and $B_{bulges}$ simulates bulges. A visual example of these augmentations is shown in Figure~\ref{fig_augmentations}.

We simulate fabric creases $B_{creases}$ by generating $N \in [5, 20]$ random linear segments. For each segment, the orientation is sampled from $\mathcal{U}(0, 2\pi)$ and the length from $\mathcal{U}(H/12, H/2)$, where $H$ is the image height. Each segment is assigned a randomized thickness of $1-4$ pixels and blurred using a Gaussian kernel with $\sigma \in [F/20, F/5]$, where $F=100$ is a spatial scale factor. Folds are assigned a random polarity $p \in \{-1, 1\}$ and scaled to a maximum displacement of $30$\,mm.

To mimic the effect of bunched blankets or hospital gowns $B_{bulges}$, we generate multi-scale, low-frequency noise clusters. This is achieved by sampling random noise at three spatial scales $\{0.5F, F, 2.0F\}$. For each scale, a coarse noise grid is generated and then expanded to the full image resolution using cubic interpolation. Each upscaled layer is refined with a Gaussian smoothing kernel where the standard deviation is scaled to the specific layer's frequency ($\sigma = F/10$). This process produces a continuous, bias field that simulates bulky surface irregularities. The integrated cluster bias is scaled to a maximum displacement of $30$\,mm.

The augmentation process ensures that the synthetic training data exhibits the non-rigid, low-frequency spatial variations characteristic of clothes and blankets, forcing the network to prioritize global anatomical proportions over local surface textures.



\subsection{Generation of Volumetric Ground Truth Labels}
\label{labels_preparation}

To generate anatomical reference labels, we utilized TotalSegmentatorMRI \citep{d2024totalsegmentator}, a robust deep learning framework capable of segmenting 56 distinct anatomical structures in whole-body MRI. To ensure the reliability of these automated segmentations as ground-truth data, we implemented a post-processing pipeline designed to improve topological consistency and eliminate segmentation artifacts. Specifically, we applied morphological hole filling to ensure anatomical completeness and 3D connected component analysis (CCA) to remove small isolated, spurious voxel clusters and misclassified regions. 

Due to the significant memory requirements of the 3D convolutional decoder, where increasing the number of target labels expands the channel dimension of the output tensor, we refined the label set to 41 target structures (Table~\ref{tab:organ_labels}). This selection was based on a dual-criteria approach: (1) a qualitative visual assessment of the automated segmentation accuracy across the cohort, and (2) the clinical relevance of the structure for automated scan planning and patient positioning. Structures with lower segmentation reliability or minimal impact on clinical workflow were excluded to optimize the model’s computational efficiency.

\begin{figure*}[t]
    \centering
    \includegraphics[width=0.8\linewidth, height=.3\textheight]{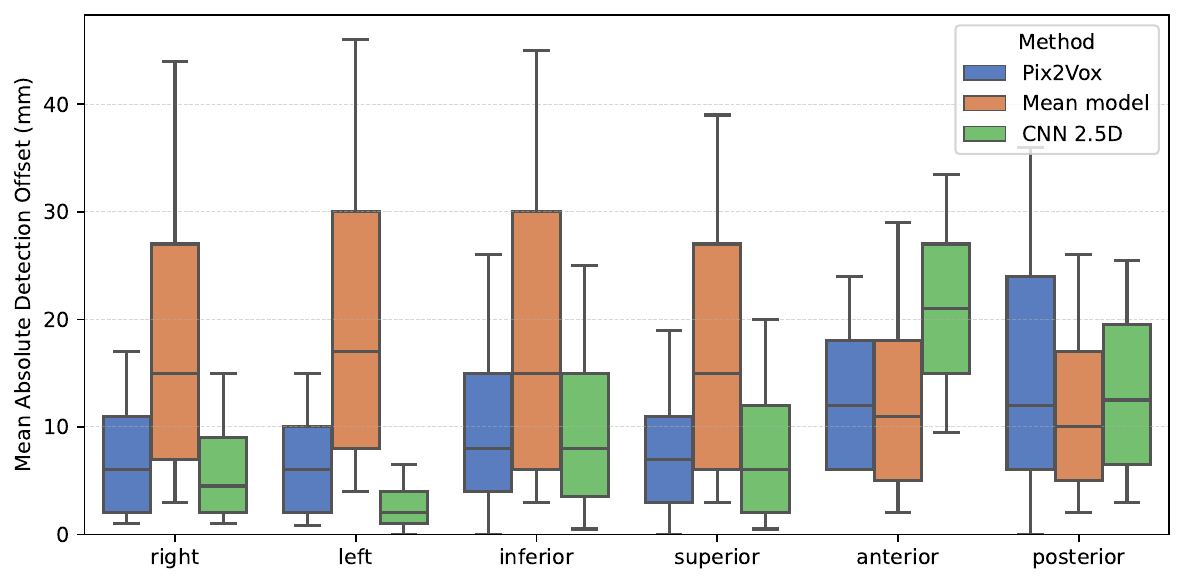}
    \caption{Mean absolute Detection Offset Error (DOE) averaged across all organs for each bounding box side (left-right, anterior-posterior, superior-inferior).}
    \label{fig_detection_error}
\end{figure*}

\subsection{Processing of Real-World Depth Images}
\label{real_depth_preparation}

To evaluate the robustness of our proposed model, we qualitatively assess its performance on real-world depth maps acquired from 11 human volunteers using a Microsoft Kinect v2 sensor (Section~\ref{real_evaluation}). To ensure consistency with the synthetic training data, the raw perspective depth frames are transformed into an orthographic projection. Specifically, raw depth frames $\mathcal{Z} \in \mathbb{R}^{U \times V}$ are first processed using a spatial Region-of-Interest (ROI) filter to isolate the subject from the scanner environment. This filter retains voxels within a valid depth range of $[1000, 2000]$ and a lateral bounding box defined by $u \in [450, 1050]$ and $v \in [400, 700]$. Utilizing the intrinsic parameters of the depth sensor and extrinsic calibration parameters, the depth coordinates are transformed into the coordinate system of the scanner, yielding a sparse point cloud $\mathbf{P} = \{x, y, z\}$. The final anterior orthographic depth image is generated by raycasting $\mathbf{P}$ onto a 2D grid. To account for no-data pixels we employ linear interpolation to reconstruct a continuous surface. The resulting depth values are min-max normalized and contrast-inverted:
\begin{equation}
    D_{real} = 1.0 - \frac{Z - Z_{min}}{Z_{max} - Z_{min}}
\end{equation}

This transformation ensures that the region in closest proximity to the sensor is represented by high-intensity values, maintaining distribution consistency with the synthetic training data $D_{norm}$ described in Section~\ref{data_preparation}.

\begin{table}[ht]
\centering
\caption{Selected organ labels grouped by tissue type}
\begin{tabular}{l ll}
\toprule
\textbf{Soft Tissue Organs} & \multicolumn{2}{c}{\textbf{Bones}} \\
\midrule
spleen               & scapula left     & vertebrae T12 \\
kidney right         & scapula right    & vertebrae T11 \\
kidney left          & clavicula left   & vertebrae T10 \\
liver                & clavicula right  & vertebrae T9  \\
stomach              & femur left       & vertebrae T8  \\
pancreas             & femur right      & vertebrae T7  \\
lung right           & hip left         & vertebrae T6  \\
lung left            & hip right        & vertebrae T5  \\
trachea              & sacrum           & vertebrae T4  \\
thyroid gland        & vertebrae L5     & vertebrae T3  \\
duodenum             & vertebrae L4     & vertebrae T2  \\
urinary bladder      & vertebrae L3     & vertebrae T1  \\
aorta                & vertebrae L2     &   \\
heart                & vertebrae L1     &   \\                     
\bottomrule
\end{tabular}
\label{tab:organ_labels}
\end{table}

\subsection{Pix2Vox Model}
\label{pix2vox}

For training, we employ a U-shaped convolutional neural network architecture that takes as input 2D depth images representing the patient's body surface and outputs 3D volumetric segmentation masks of internal anatomical structures (Figure~\ref{fig_model_arch}). The network is designed to bridge the dimensional gap between 2D input and 3D output using a combination of standard 2D encoder and 3D decoder modules, connected via custom 2D-to-3D conversion layers placed at the bottleneck and each skip connection.

The encoder consists of a series of convolutional and pooling layers that extract increasingly abstract spatial features from the input depth image. These features are encoded as 2D feature maps of shape $B \times C_{2D} \times H \times W$, where $B$ denotes the batch size, $C_{2D}$ the number of channels, and $H$, $W$ the spatial height and width, respectively.

To transition from 2D to 3D, we introduce conversion layers that reshape and transform the encoded 2D features into a format suitable for the 3D decoder. Specifically, for a given layer $l$, we apply the following transformation:
\begin{enumerate}
    \item First, we unsqueeze the 2D feature maps along a new feature axis, producing a 5D tensor of shape $B \times 1 \times C_{2D} \times H \times W$. This tensor is interpreted as a volumetric feature map with one feature map and a depth dimension equal to $C_{2D}$.
    \item Next, we apply a 3D convolutional layer with kernel size $k_l \times 1 \times 1$, stride $s_l \times 1 \times 1$, and output feature size $f_l$. The $k_l$, $s_l$, and $f_l$ parameters are chosen individually for each layer $l$ to produce an appropriate depth dimension $D$  and number of features $C_{3D}$ for input into the 3D decoder (Figure~\ref{fig_model_arch}).
\end{enumerate}

The decoder is composed of standard 3D convolutional layers with upsampling operations to progressively reconstruct the full volumetric segmentation output. The decoder also incorporates skip connections from the encoder, each of which passes through its corresponding 2D-to-3D conversion layer to ensure dimensional compatibility.

Prior to network processing, the input depth map is resized to $256 \times 256$ pixels. The final output volume has a dimension of $42 \times 64 \times 256 \times 256$, corresponding to $C_{3D} \times D \times H \times W$. The architecture employs a 5-layer encoder and a matching 5-layer decoder. Specifically, the number of channels $C_{2D}$ across the 2D encoder layers is configured as 64, 64, 128, 256, and 512 from high-resolution to low-resolution features. Correspondingly, the number of feature channels $C_{3D}$ within the 3D decoder layers is configured as 16, 16, 32, 64, and 128. To ensure dimensional compatibility, the skip connections utilize kernel sizes $k_l$ of $(1, 1, 1)$, $(2, 1, 1)$, $(8, 1, 1)$, $(32, 1, 1)$, and $(128, 1, 1)$ with matching strides $s_l$ of $(1, 1, 1)$, $(2, 1, 1)$, $(8, 1, 1)$, $(32, 1, 1)$, and $(128, 1, 1)$, effectively bridging the dimensions between the 2D encoder and the 3D decoder.

This hybrid 2D-3D design enables efficient processing of the 2D input depth images while producing dense 3D predictions, balancing memory efficiency with spatial expressiveness.

\section{Experiments and Results}

\begin{figure*}[t]
    \centering
    \includegraphics[width=0.8\linewidth, height=.3\textheight]{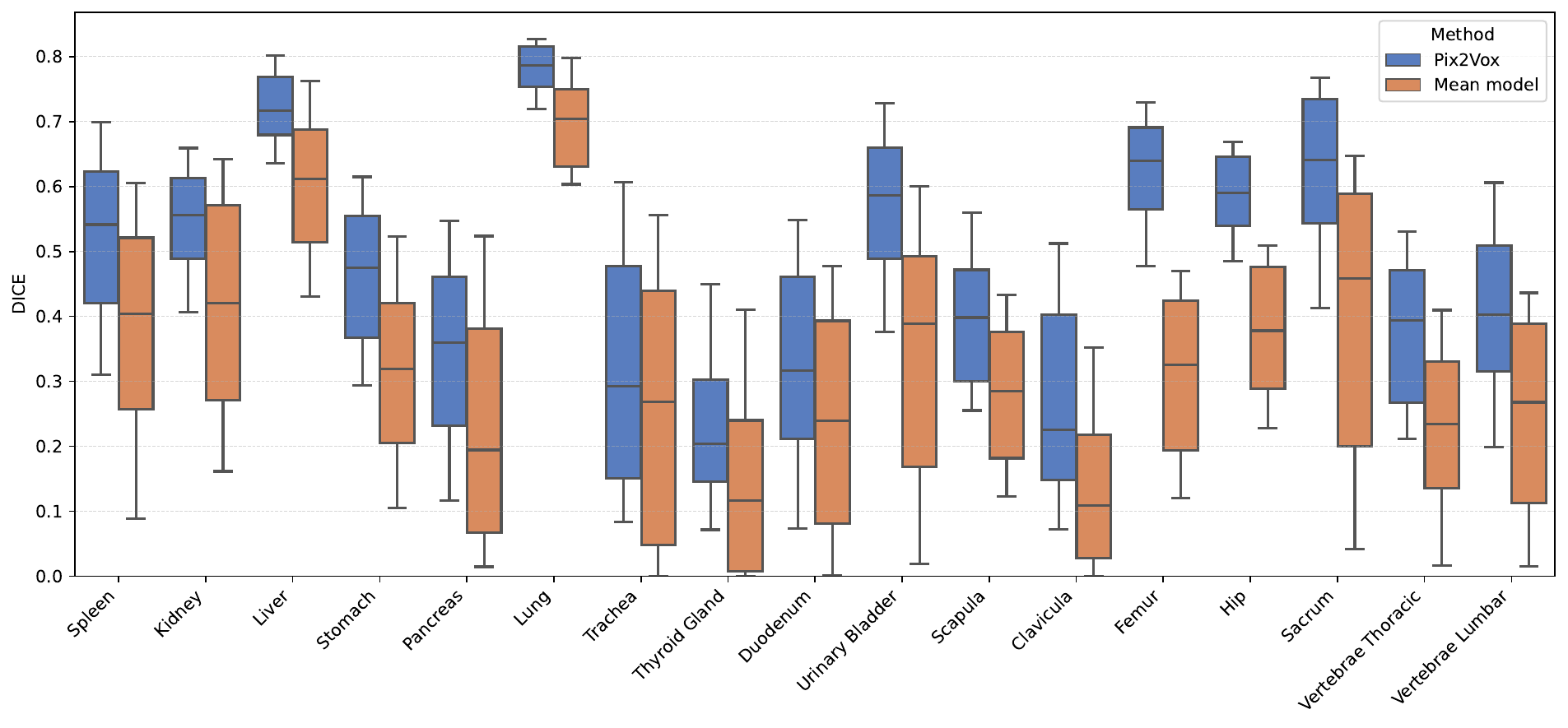}
    \caption{Dice coefficient for volumetric organs segmentation. Values are averaged for bilateral organs (e.g., kidneys, lungs) and vertebral subgroups (thoracic and lumbar vertebrae).}
    \label{fig_dice}
\end{figure*}

To train our model, we randomly selected 8,528 pairs of depth images and their corresponding segmentation masks from the preprocessed dataset ( Section~\ref{data_preparation}). A separate validation set of 494 samples was used to monitor training performance and guide model selection and hyperparameter tuning. For evaluation, we employed a dedicated test set of 998 samples, which was held out during training. This dataset size allows for a quantitative assessment of the model's performance across a range of anatomical variations.

\subsection{Training and evaluation details}

We train our model in a multi-label segmentation setting, where the network outputs a separate channel for each of the 41 anatomical structures. During training, a sigmoid activation is applied channel-wise to compute a combined Dice and binary cross-entropy loss, with both components equally weighted at 0.5. To enhance the model's robustness to anatomical variability and patient positioning, we apply data augmentations including random shifts, scaling, and rotations during training. This encourages the model to generalize across a broader range of spatial configurations. Training is performed using the Adam optimizer with an initial learning rate of 0.001. We apply a linear warm-up strategy over the first 1,000 training steps to stabilize early training, followed by cosine annealing to gradually reduce the learning rate to zero over the remaining steps. We use a batch size of 8, and training is carried out for 15,000 steps in total. To further improve robustness and reduce prediction noise during evaluation, we apply a post-processing step based on connected component analysis. This removes small, isolated predictions that are spatially distant from the main predicted regions, thereby reducing the likelihood of large errors caused by spurious or misclassified voxels.

 \begin{figure*}[t]
    \centering
    \includegraphics[width=0.8\linewidth, height=.3\textheight]{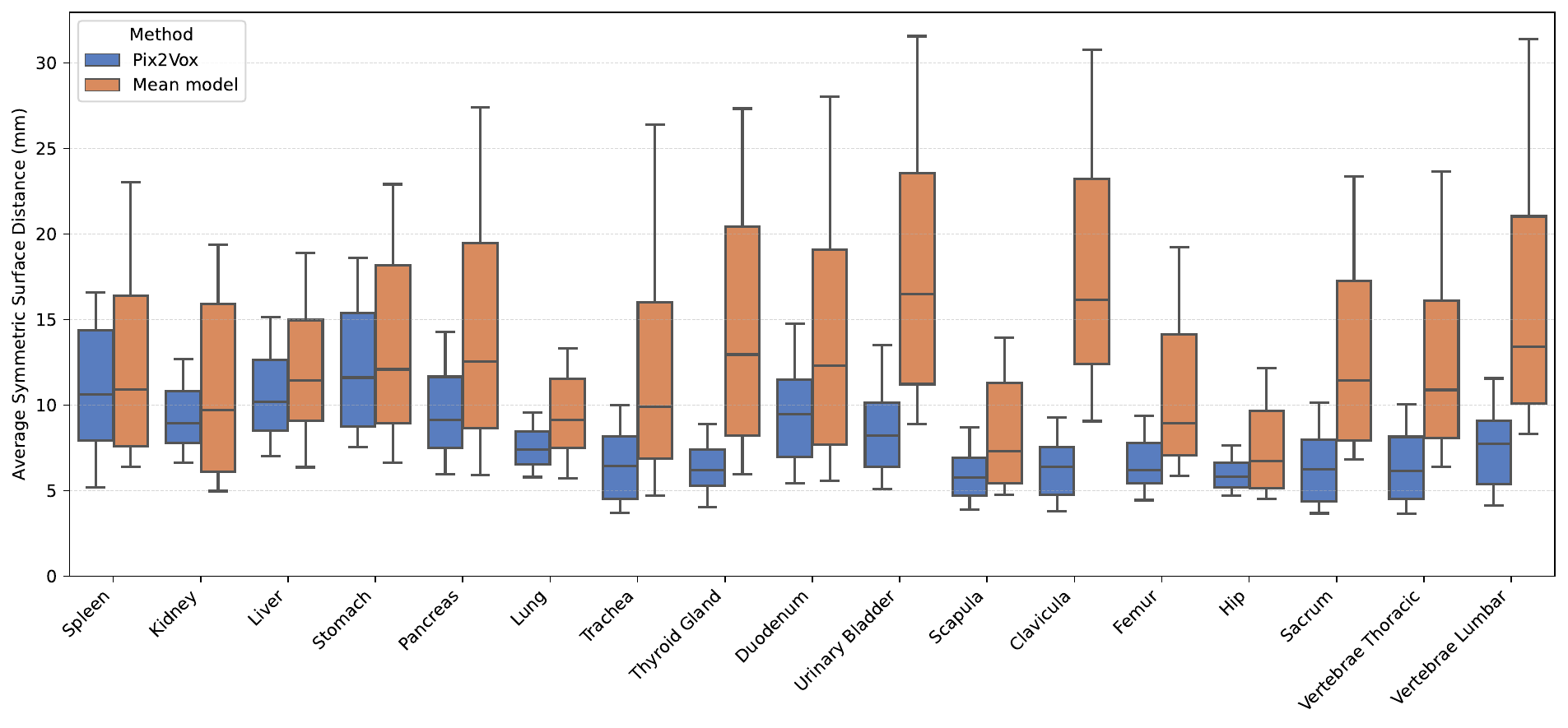}
    \caption{Average Symmetric Surface Distance (ASSD) for volumetric organs segmentation. Values are averaged for bilateral organs (e.g., kidneys, lungs) and vertebral subgroups (thoracic and lumbar vertebrae).}
    \label{fig_sd}
\end{figure*}

\subsection{Baselines}

\paragraph{Mean model.} Given the well-aligned patient positioning in the NAKO MRI dataset and the consistent anatomical layout across subjects, we exploit this property to construct a mean segmentation baseline model. Despite variability in patient-specific characteristics such as age, height, and body shape, the high degree of spatial alignment in the dataset enables the generation of an average anatomical representation from the training data. To build this baseline, we aggregate all ground truth segmentations from the training set by computing the voxel-wise mean across all examples. Specifically, each label channel is summed across all training subjects, divided by the total number of samples, and then thresholded at 50\% of the maximum voxel intensity for that label to obtain a binary segmentation. This process yields a static, patient-agnostic model that serves as a reference for evaluating the effectiveness of learning-based methods. Although simplistic, this mean model provides a strong benchmark in settings with highly standardized acquisition protocols, offering insights into how much a learning-based model improves upon pure spatial prior knowledge.

\paragraph{2.5D CNN.} Inspired by prior work by \cite{geissler2025predicting} and \cite{kats2025internal}, which leverage 2D CNNs to predict coronal plane projections of anatomical structures, we extend this concept to a more comprehensive 2.5D setting. Specifically, we employ separate 2D CNN models to predict the organ projections onto all three principal anatomical planes: coronal, sagittal, and axial. Each model is trained independently to generate a binary mask representing the projection of target anatomical structures in its respective plane, using the same depth image as input. By aggregating the outputs of these three models, we can infer 3D bounding boxes for each organ by combining the spatial extents across all dimensions. This approach offers a computationally efficient alternative to full 3D segmentation, while still capturing spatial context through multi-planar analysis.

 \begin{figure*}[thbp]
    \centering
    \includegraphics[width=0.98\linewidth, height=.40\textheight]{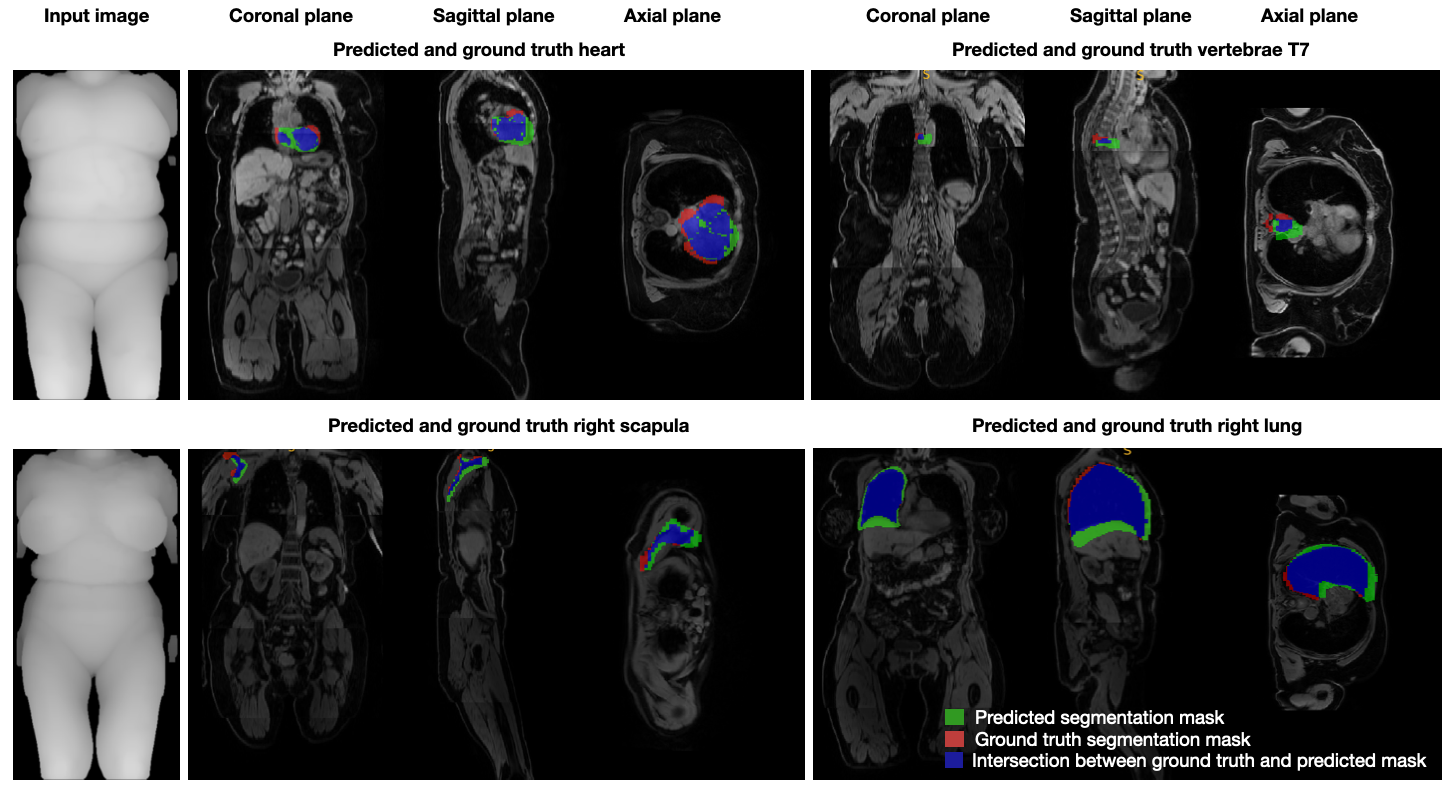}
    \caption{Qualitative comparison between predicted and ground truth volumetric segmentations.}
    \label{fig_qualitative}
\end{figure*}

\subsection{Metrics}

We evaluate the localization and segmentation performance of all models using a combination of complementary metrics that capture both spatial accuracy and anatomical shape fidelity. Bounding boxes are derived from the volumetric segmentation masks produced by the mean model, the Pix2Vox model, and the three-plane predictions of the 2.5D CNN. To quantify localization accuracy, we calculate the Mean Absolute Detection Offset Error (DOE), defined as the absolute distance between the corresponding faces of the predicted and ground truth bounding boxes along each axis:
\begin{equation}
    \text{DOE} = \frac{1}{N} \sum_{i=1}^{N} | \text{face}_{i, \text{pred}} - \text{face}_{i, \text{gt}} |
\end{equation}
where $\text{face} \in \{x_{min}, x_{max}, y_{min}, y_{max}, z_{min}, z_{max}\}$ represents the coordinates of the six bounding box planes.
To better understand model behavior in different spatial dimensions, DOE is reported separately for each side of the bounding box (e.g., left, right, anterior, posterior, superior, inferior). This directional breakdown allows us to assess how accurately each method positions the anatomical structures along the full 3D extent. Importantly, this metric reflects practical constraints in clinical imaging scenarios, particularly in automated patient positioning and scan planning, where under- or over-estimating organ boundaries can lead to suboptimal coverage or increased scan time. To assess the model's ability to reconstruct organ shape, we compute the Average Symmetric Surface Distance (ASSD) between the predicted and ground truth segmentations, which captures discrepancies at the organ boundary. In addition, we report the Dice coefficient as a standard volumetric overlap metric. Both Dice and ASSD are calculated only for models that produce full 3D volumetric predictions, enabling us to evaluate how well the organ shape can be recovered from 2D depth image inputs alone.

\subsection{Results}

\subsubsection{Localization Performance}
\label{sec:results_localization}

Quantitative analysis using DOE highlights the balanced 3D localization performance of the proposed Pix2Vox model across all anatomical axes, as illustrated in Figure~\ref{fig_detection_error}. Within the coronal plane, Pix2Vox demonstrates a substantial improvement over the mean model. Specifically, it achieves a mean absolute error of $7.87 \pm 7.69$\,mm for the right boundary and $6.87 \pm 6.62$\,mm for the left, significantly outperforming the baseline values of $19.86 \pm 17.14$\,mm and $21.23 \pm 17.19$\,mm, respectively.

The 2.5D CNN approach, leveraging multi-planar projections, achieves the highest precision in lateral localization with a left-side DOE of $2.79 \pm 2.93$\,mm. While both deep learning approaches exhibit comparable performance along the superior-inferior axis, with errors ranging between $8.39$\,mm and $11.36$\,mm, the anterior-posterior axis remains the most challenging dimension for the 2.5D method. In the anterior direction, Pix2Vox yields an error of $13.29 \pm 8.41$\,mm, which is notably superior to the $21.31 \pm 9.53$\,mm error produced by the 2.5D CNN. Interestingly, the mean model provides a competitive baseline along the depth axis, yielding errors of $13.00 \pm 10.63$\,mm in the anterior direction and $12.90 \pm 12.48$\,mm in the posterior direction.

Furthermore, it is worth noting that while Pix2Vox consistently outperforms the baselines in average metrics and across the majority of soft-tissue organs and large skeletal structures, the 2.5D CNN model provides better results across individual vertebral segments (Appendix~(\ref{app_metrics})).

\begin{figure*}[thbp]
    \centering
    \includegraphics[width=0.98\linewidth]{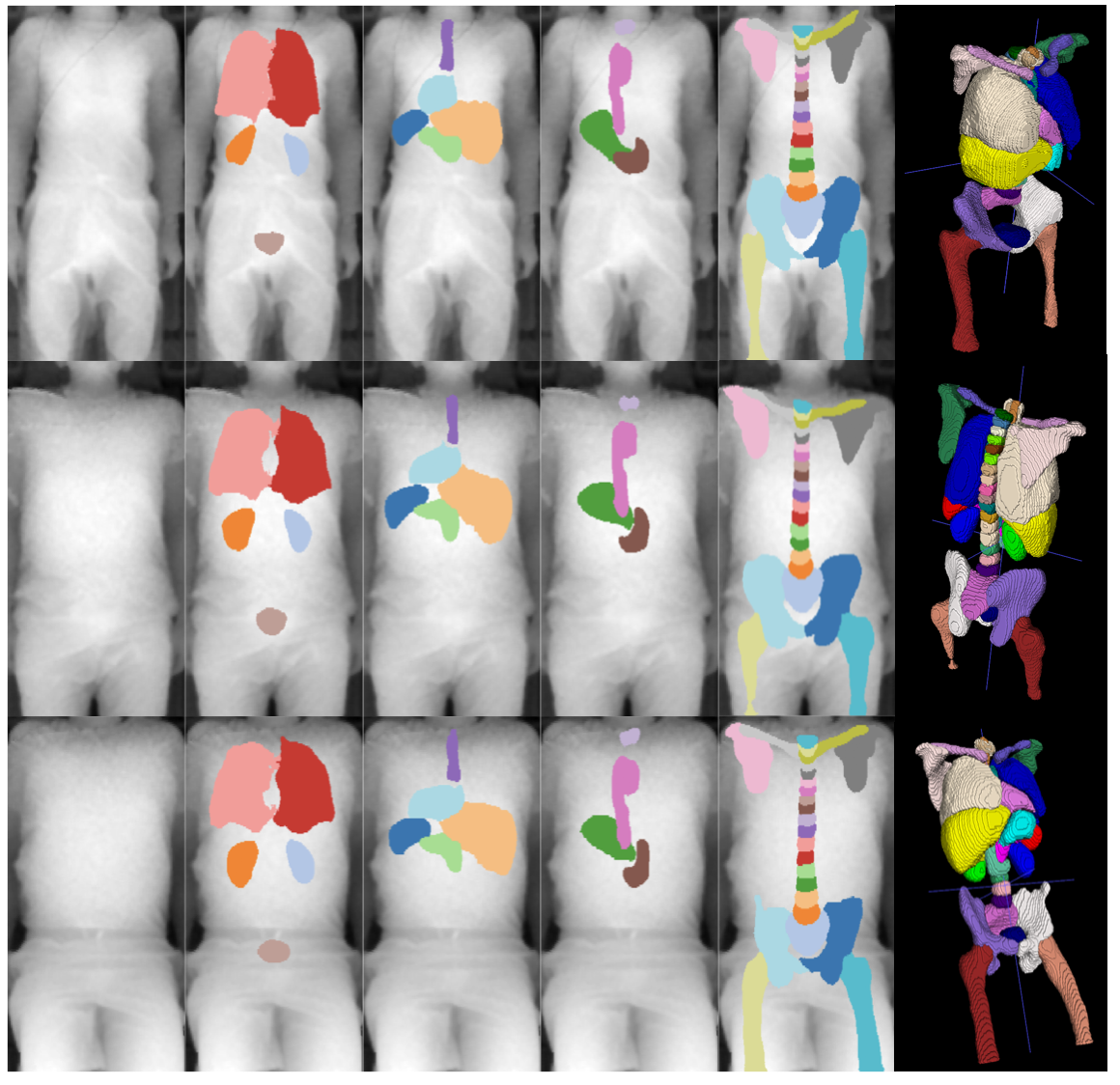}
    \caption{Qualitative evaluation on real-world depth images of volunteers positioned in an MRI scanner, following the setup described in Section~\ref{sec_method}. The predicted internal structures are displayed as volumetric segmentations projected onto the coronal plane. From left to right: (1) input depth image; (2) overlay of lungs, kidneys, and urinary bladder; (3) overlay of trachea, heart, spleen, liver, and pancreas; (4) overlay of thyroid gland, aorta, stomach, and duodenum; (5) overlay of skeletal system; and (6) volumetric segmentation, shown in a different perspective for each subject.}
    \label{fig_realworld}
\end{figure*}

\subsubsection{Segmentation and Shape Fidelity}
\label{sec:results_segmentation}

The volumetric reconstruction performance was further evaluated using the Dice similarity (Figure~\ref{fig_dice}) coefficient and ASSD (Figure~\ref{fig_sd}). A detailed per-organ breakdown for all 41 anatomical structures is provided in Appendix~\ref{app_metrics}. The Pix2Vox model achieved an overall mean Dice score of $0.48 \pm 0.20$, representing a clear advancement over the mean model baseline of $0.34 \pm 0.21$. Anatomical analysis reveals that large organs with a high correlation to the body surface exhibit the most robust reconstruction. Specifically, the lungs and liver achieved Dice scores of $0.78 \pm 0.04$ and $0.71 \pm 0.07$, respectively. In contrast, smaller or highly variable structures, such as the thyroid gland at $0.24 \pm 0.15$ and the trachea at $0.32 \pm 0.19$, presented greater challenges for surface-based volumetric estimation. Skeletal structures also showed competitive performance, with the femur at $0.62 \pm 0.10$ and the sacrum at $0.61 \pm 0.15$ yielding stable overlap metrics.

The surface fidelity metrics follow a similar trend. Pix2Vox reduced the overall mean ASSD to $8.34 \pm 8.11$\,mm, compared to $13.17 \pm 10.84$\,mm for the mean model. The highest surface accuracy was observed for the scapula and the hip, with errors of $5.96 \pm 1.65$\,mm and $6.08 \pm 1.27$\,mm, respectively. Notably, while some internal organs exhibit lower Dice scores due to their small volume, the corresponding ASSD values remain relatively low, as seen with the $6.58$\,mm error for the thyroid gland.

\subsubsection{Evaluation on Real-World Depth Images}
\label{real_evaluation}

To assess the translational potential of our framework, we evaluated the trained model on depth maps acquired from volunteers using a Microsoft Kinect v2 sensor. As illustrated in Figure~\ref{fig_realworld}, the model reconstructs internal anatomical structures that appear spatially consistent with the subject's external surface manifold. The predicted organ locations and global orientations align with expected human anatomy, suggesting that the model is robust to the real-world depth images.

However, detailed visual inspection reveals specific reconstruction artifacts inherent to surface-only volumetric estimation. In the skeletal overlays, particularly in the middle image of Figure~\ref{fig_realworld}, morphological artifacts are noticeable in the femoral regions, appearing as unusually narrowing towards bottom of the image. Furthermore, the lung segmentations in the middle and bottom rows exhibit localized peak-like artifacts. While the overall abdominal arrangement is preserved, the morphology of the liver in middle image shows a loss of characteristic lobar sharpness.

\section{Discussion}
\label{sec:discussion}

The quantitative and qualitative results indicate that 3D internal anatomy can be estimated from 2D surface depth information, though the degree of accuracy is contingent upon the specific organ of interest and the anatomical axis.

Analysis of the Detection Offset Error reveals distinct trade-offs between the evaluated architectures. While the Pix2Vox model provided more balanced 3D localization across all axes, the 2.5D CNN achieved the highest precision in the lateral left-right dimensions. This is likely attributable to its multi-planar projection approach, which excels at identifying boundary edges in 2D space. However, the performance degradation of the 2.5D CNN along the anterior-posterior axis suggests that the absence of a dedicated 3D decoding mechanism limits its ability to resolve depth-specific structural cues. The lower performance in the anterior-posterior direction observed for both deep learning approaches is not unexpected, as a single-view coronal depth map provides limited information regarding the posterior extent of internal organs.

The disparity in Dice scores between large organs, such as the lungs, and smaller structures, such as the thyroid gland, suggests that volumetric overlap is highly dependent on the absolute size of the target. However, the relatively low ASSD for small organs indicates that the model accurately localizes the anatomical center of these structures even when it cannot precisely resolve their morphological boundaries.

\subsection{Real-World Generalization and Artifacts}

The evaluation on real-world Kinect v2 data (Section~\ref{real_evaluation}) demonstrates that the model is robust to real-world depth acquisition, but it also highlights inherent limitations in surface-to-volume inference. The narrowing artifacts observed in the femoral regions and the peak-like artifacts in the lung segmentations suggest a sensitivity to surface curvature and sensor noise. The loss of lobar sharpness in the liver morphology further indicates that internal soft-tissue boundaries are not fully constrained by the surface manifold. These discrepancies suggest that while global positioning is effectively captured, fine-grained boundary reconstruction of non-surface-correlated tissues remains a challenge.

\subsection{Limitations and Clinical Considerations}

Several technical constraints must be addressed for clinical implementation:
\begin{itemize}
    \item \textbf{Domain Gap:} The depth images used in this study were synthetically generated from MRI-derived segmentations and surface reconstructions, which may not fully capture the variability present in real-world depth camera outputs. In clinical environments, additional sources of noise, such as occlusions, limited spatial resolution, and sensor-specific distortions, can affect image quality. Furthermore, patients may be covered with thick blankets or use assistive devices that obscure key body surface features. Bridging this domain gap is crucial for real-world applicability and may require domain adaptation techniques or model fine-tuning using real sensor-acquired data.
    \item \textbf{Positional Constraints:} The training data used in this work consisted of patients in a standardized supine position with minimal variation in posture. Although geometric data augmentations were employed during training to increase robustness, real-world scenarios often involve a wider range of patient poses due to comfort needs or mobility restrictions. The current model may be sensitive to these variations, and further research is needed to improve its generalizability to diverse body orientations.
\end{itemize}

The primary clinical motivation for this 3D localization task is to enable accurate, automated patient table positioning in imaging workflows. For standard clinical protocols, an alignment accuracy within a threshold of 10 to 20,mm is typically considered sufficient to ensure that the target anatomy resides entirely within the scanner's optimal field of view~\citep{booij2019accuracy, saltybaeva2017vertical}. By achieving a mean absolute DOE within this target range, the proposed Pix2Vox model satisfies these clinical requirements. These results indicate the technical feasibility of utilizing depth-based volumetric estimation for automated patient table positioning within the scanner bore.  


\acks{We gratefully acknowledge the financial support by German
Research Foundation: DFG, HE 7364/10-1, project number
500498869.}

%
\ethics{The work follows appropriate ethical standards in conducting research and writing the manuscript, following all applicable laws and regulations regarding treatment of human subjects.}

\coi{We declare we do not have conflicts of interest.}

\data{Data from the German National Cohort (NAKO) are available for research purposes upon registration and formal request through the NAKO Research Platform: \url{https://nako.de/en/research/}. Access is subject to approval by the NAKO Data Use and Access Committee and compliance with data protection regulations.}

\bibliography{melba}

@article{van2020cinderellas,
  title={The Cinderellas of the scanner: Magnetic resonance imaging'pre-scan'and'post-scan'times: Their determinants and impact on patient throughput},
  author={Van Rooyen, Marthinus B and Pitcher, Richard D},
  journal={SA Journal of Radiology},
  volume={24},
  number={1},
  pages={1--6},
  year={2020},
  publisher={AOSIS}
}

@article{koken2009towards,
  title={Towards automatic patient positioning and scan planning using continuously moving table MR imaging},
  author={Koken, Peter and Dries, Sebastian PM and Keupp, Jochen and Bystrov, Daniel and Pekar, Vladimir and B{\"o}rnert, Peter},
  journal={Magnetic Resonance in Medicine: An Official Journal of the International Society for Magnetic Resonance in Medicine},
  volume={62},
  number={4},
  pages={1067--1072},
  year={2009},
  publisher={Wiley Online Library}
}

@article{incetan2020rgb,
  title={RGB-D camera-based clinical workflow optimization for rotational angiography},
  author={{\.I}ncetan, Ka{\u{g}}an and Mohan, Rishi and Stoutjesdijk, Henry and Fernandes, Nelson and de Jager, Bram},
  journal={IEEE Sensors Journal},
  volume={20},
  number={15},
  pages={8867--8874},
  year={2020},
  publisher={IEEE}
}

@article{karanam2020towards,
  title={Towards contactless patient positioning},
  author={Karanam, Srikrishna and Li, Ren and Yang, Fan and Hu, Wei and Chen, Terrence and Wu, Ziyan},
  journal={IEEE transactions on medical imaging},
  volume={39},
  number={8},
  pages={2701--2710},
  year={2020},
  publisher={IEEE}
}

@article{shetty2023boss,
  title={BOSS: Bones, organs and skin shape model},
  author={Shetty, Karthik and Birkhold, Annette and Jaganathan, Srikrishna and Strobel, Norbert and Egger, Bernhard and Kowarschik, Markus and Maier, Andreas},
  journal={Computers in Biology and Medicine},
  volume={165},
  pages={107383},
  year={2023},
  publisher={Elsevier}
}

@article{loper2015smpl,
  title={SMPL: a skinned multi-person linear model},
  author={Loper, Matthew and Mahmood, Naureen and Romero, Javier and Pons-Moll, Gerard and Black, Michael J},
  journal={ACM Transactions on Graphics (TOG)},
  volume={34},
  number={6},
  pages={1--16},
  year={2015},
  publisher={ACM New York, NY, USA}
}

@article{bamberg2022whole,
  title={Whole-body Magnetic Resonance Imaging in the German National Cohort (NAKO): Design \& Current Status},
  author={Bamberg, F},
  journal={European Journal of Public Health},
  volume={32},
  number={Supplement\_3},
  pages={ckac129--029},
  year={2022},
  publisher={Oxford University Press}
}

@article{d2024totalsegmentator,
  title={TotalSegmentator MRI: Sequence-Independent Segmentation of 59 Anatomical Structures in MR images},
  author={D'Antonoli, Tugba Akinci and Berger, Lucas K and Indrakanti, Ashraya K and Vishwanathan, Nathan and Wei{\ss}, Jakob and Jung, Matthias and Berkarda, Zeynep and Rau, Alexander and Reisert, Marco and K{\"u}stner, Thomas and others},
  journal={arXiv preprint arXiv:2405.19492},
  year={2024}
}

@article{danilouchkine2005operator,
  title={Operator induced variability in cardiovascular MR: left ventricular measurements and their reproducibility},
  author={Danilouchkine, Mikhail G and Westenberg, Jos JM and de Roos, Albert and Reiber, Johan HC and Lelieveldt, Boudewijn PF},
  journal={Journal of Cardiovascular Magnetic Resonance},
  volume={7},
  number={2},
  pages={447--457},
  year={2005},
  publisher={Taylor \& Francis}
}

@article{shafieizargar2023systematic,
  title={Systematic review of reconstruction techniques for accelerated quantitative MRI},
  author={Shafieizargar, Banafshe and Byanju, Riwaj and Sijbers, Jan and Klein, Stefan and den Dekker, Arnold J and Poot, Dirk HJ},
  journal={Magnetic Resonance in Medicine},
  volume={90},
  number={3},
  pages={1172--1208},
  year={2023},
  publisher={Wiley Online Library}
}

@article{heckel2024deep,
  title={Deep learning for accelerated and robust MRI reconstruction},
  author={Heckel, Reinhard and Jacob, Mathews and Chaudhari, Akshay and Perlman, Or and Shimron, Efrat},
  journal={Magnetic Resonance Materials in Physics, Biology and Medicine},
  volume={37},
  number={3},
  pages={335--368},
  year={2024},
  publisher={Springer}
}

@article{booij2021automated,
  title={Automated patient positioning in CT using a 3D camera for body contour detection: accuracy in pediatric patients},
  author={Booij, Ronald and van Straten, Marcel and Wimmer, Andreas and Budde, Ricardo PJ},
  journal={European radiology},
  volume={31},
  number={1},
  pages={131--138},
  year={2021},
  publisher={Springer}
}

@article{booij2019accuracy,
  title={Accuracy of automated patient positioning in CT using a 3D camera for body contour detection},
  author={Booij, Ronald and Budde, Ricardo PJ and Dijkshoorn, Marcel L and van Straten, Marcel},
  journal={European Radiology},
  volume={29},
  number={4},
  pages={2079--2088},
  year={2019},
  publisher={Springer}
}

@article{henrich2025looc,
  title={Looc: Localizing organs using occupancy networks and body surface depth images},
  author={Henrich, Pit and Mathis-Ullrich, Franziska},
  journal={IEEE Access},
  year={2025},
  publisher={IEEE}
}

@article{streit2021analysis,
  title={Analysis of core processes of the MRI workflow for improved capacity utilization},
  author={Streit, Ulrike and Uhlig, Johannes and Lotz, Joachim and Panahi, Babak and Hosseini, Ali Seif Amir},
  journal={European Journal of Radiology},
  volume={138},
  pages={109648},
  year={2021},
  publisher={Elsevier}
}

@article{saltybaeva2017vertical,
  title={Vertical off-centering affects organ dose in chest CT: evidence from Monte Carlo simulations in anthropomorphic phantoms},
  author={Saltybaeva, Natalia and Alkadhi, Hatem},
  journal={Medical physics},
  volume={44},
  number={11},
  pages={5697--5704},
  year={2017},
  publisher={Wiley Online Library}
}

@inproceedings{teixeira2023automated,
  title={Automated CT Lung Cancer Screening Workflow Using 3D Camera},
  author={Teixeira, Brian and Singh, Vivek and Tamersoy, Birgi and Prokein, Andreas and Kapoor, Ankur},
  booktitle={International Conference on Medical Image Computing and Computer-Assisted Intervention},
  pages={423--431},
  year={2023},
  organization={Springer}
}

@inproceedings{guo2022smpl,
  title={SMPL-A: Modeling person-specific deformable anatomy},
  author={Guo, Hengtao and Planche, Benjamin and Zheng, Meng and Karanam, Srikrishna and Chen, Terrence and Wu, Ziyan},
  booktitle={Proceedings of the IEEE/CVF Conference on Computer Vision and Pattern Recognition},
  pages={20814--20823},
  year={2022}
}

@inproceedings{keller2022osso,
  title={OSSO: Obtaining skeletal shape from outside},
  author={Keller, Marilyn and Zuffi, Silvia and Black, Michael J and Pujades, Sergi},
  booktitle={Proceedings of the IEEE/CVF conference on computer vision and pattern recognition},
  pages={20492--20501},
  year={2022}
}

@inproceedings{keller2024hit,
  title={HIT: Estimating internal human implicit tissues from the body surface},
  author={Keller, Marilyn and Arora, Vaibhav and Dakri, Abdelmouttaleb and Chandhok, Shivam and Machann, J{\"u}rgen and Fritsche, Andreas and Black, Michael J and Pujades, Sergi},
  booktitle={Proceedings of the IEEE/CVF Conference on Computer Vision and Pattern Recognition},
  pages={3480--3490},
  year={2024}
}

@inproceedings{kats2025internal,
  title={Internal Organ Localization using Depth Images: A Framework for Automated MRI Patient Positioning},
  author={Kats, Eytan and Gei{\ss}ler, Kai and Hirsch, Jochen G and Heldman, Stefan and Heinrich, Mattias P},
  booktitle={BVM Workshop},
  pages={324--329},
  year={2025},
  organization={Springer}
}

@inproceedings{geissler2025predicting,
  title={Predicting anatomical structures from 2D depth images of patients},
  author={Gei{\ss}ler, Kai and Mensing, Daniel and Kohlbrandt, Temke and Hirsch, Jochen G and Heldmann, Stefan},
  booktitle={Medical Imaging 2025: Image Processing},
  volume={13406},
  pages={407--411},
  year={2025},
  organization={SPIE}
}

\clearpage

\appendix
\section{Detailed Performance Metrics}
\label{app_metrics}

\begin{figure*}[ht] 
    \centering
    
    \includegraphics[width=0.85\linewidth]{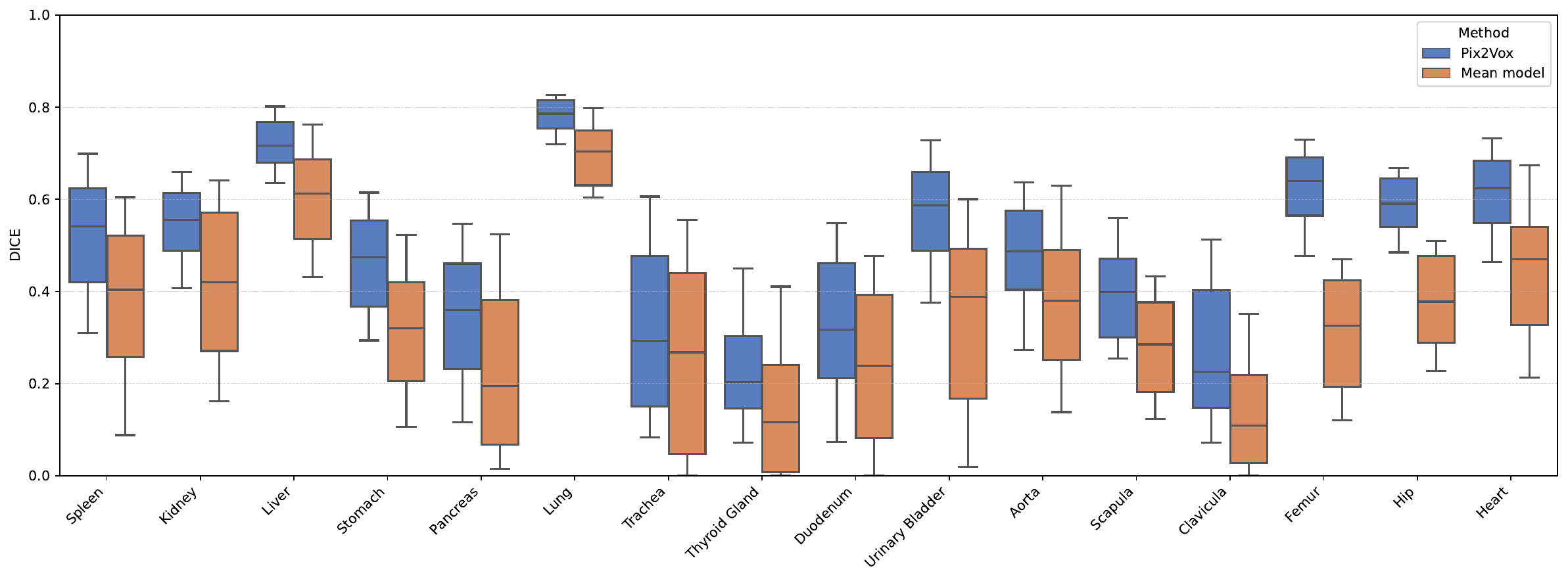}
    \caption{Dice similarity coefficient for internal organs.}
    \label{fig_dice_others}
    
    \vspace{1.5em} 
    
    \includegraphics[width=0.85\linewidth]{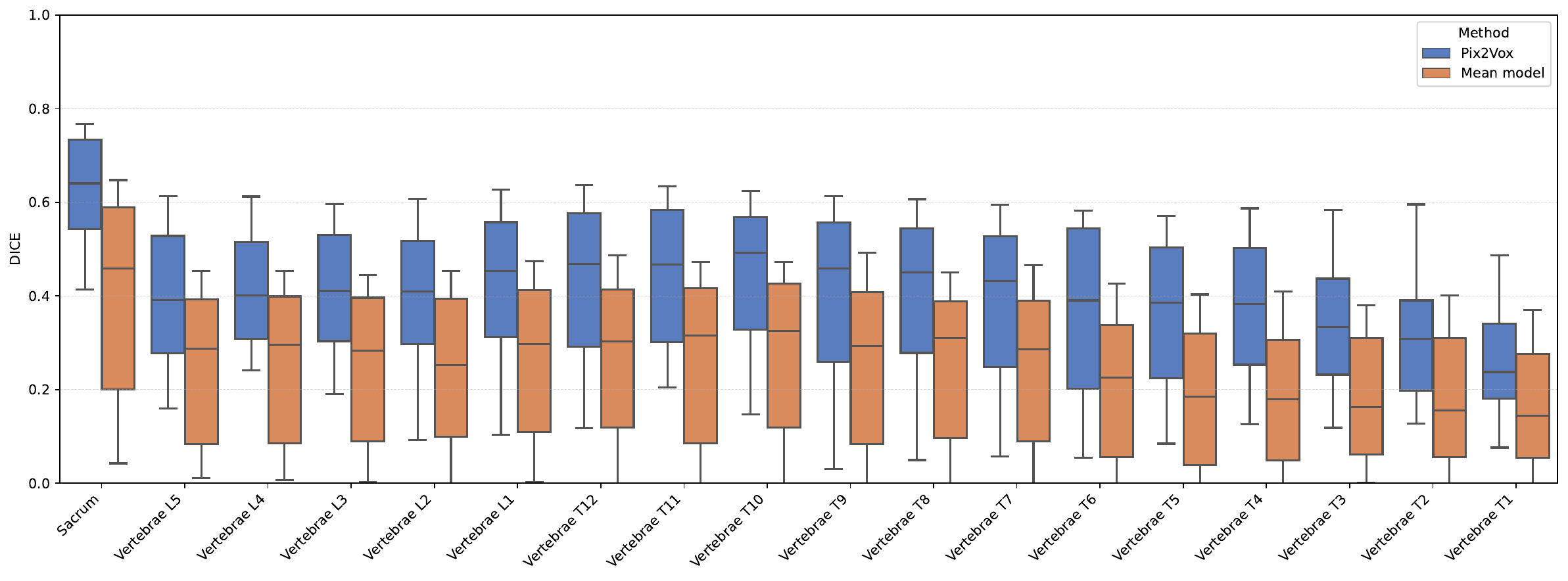}
    \caption{Dice similarity coefficient for internal organs.}
    \label{fig_dice_vertebrae}
    
    \vspace{1.5em}

    \includegraphics[width=0.85\linewidth]{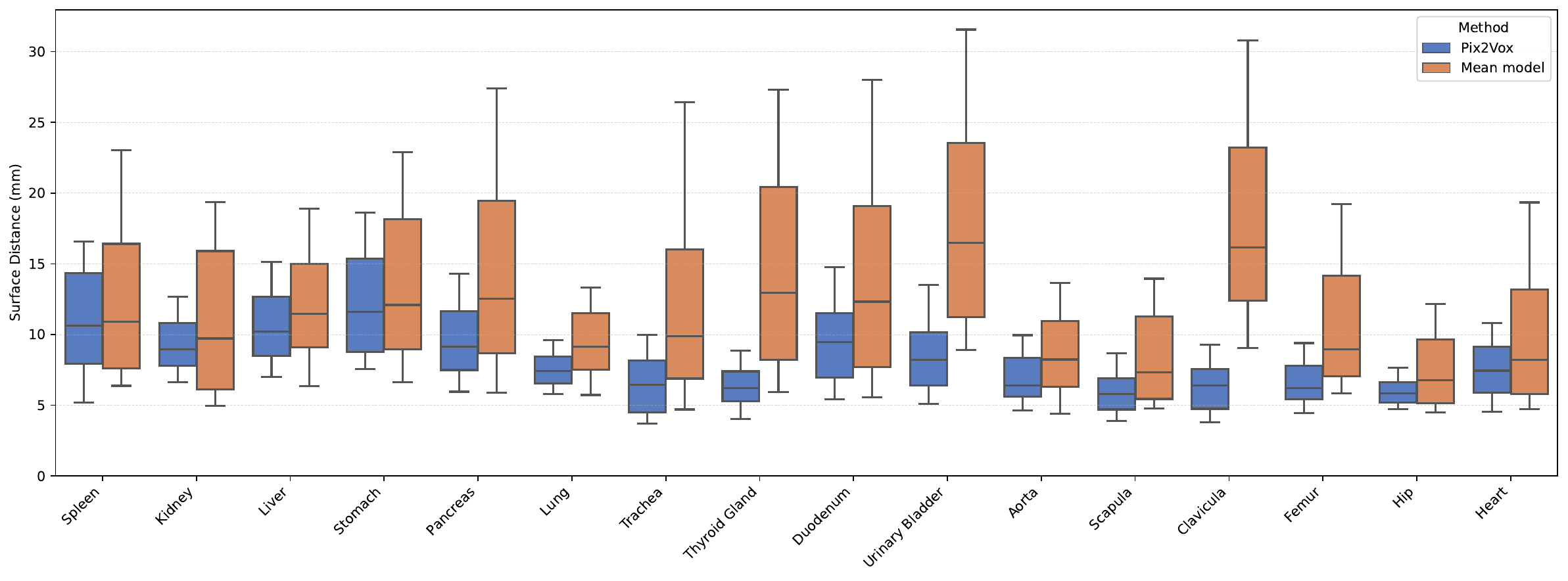}
    \caption{Surface distance metrics for internal organs.}
    \label{fig_sd_others}
\end{figure*}

\begin{figure*}[ht] 
    \centering
    
    \includegraphics[width=0.85\linewidth]{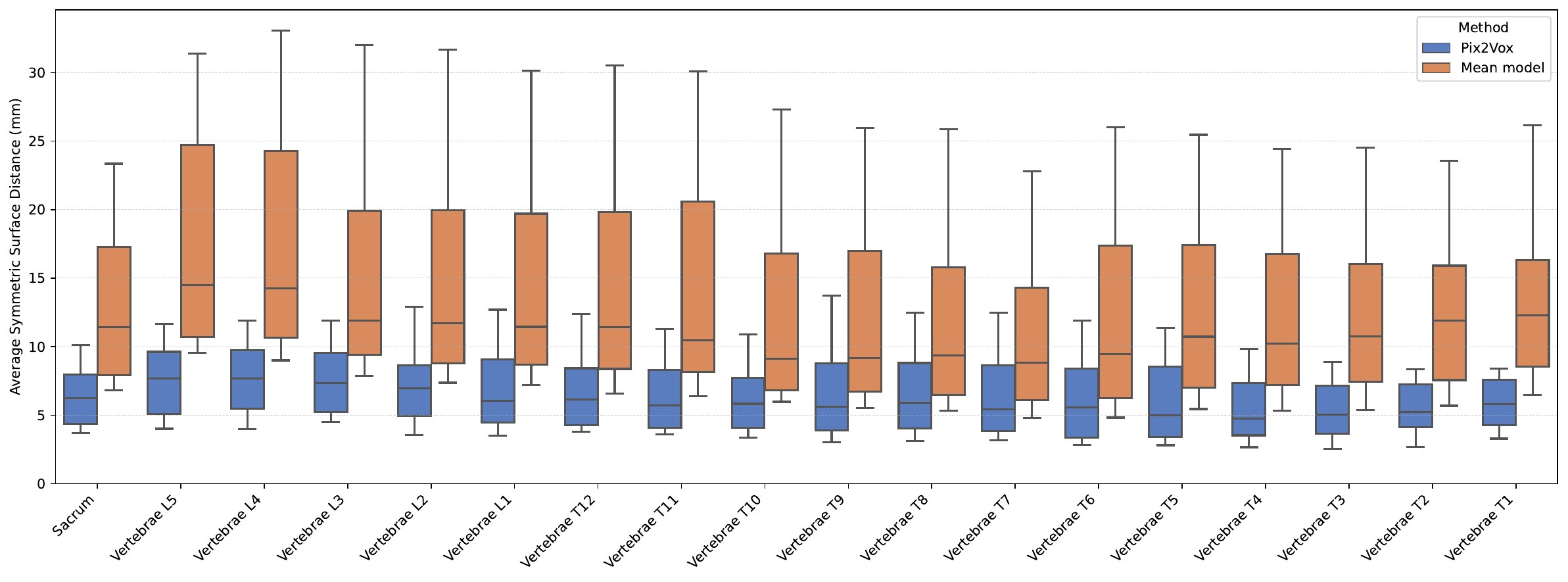}
    \caption{Surface distance metrics for internal organs.}
    \label{fig_sd_vertebrae}
    
    \vspace{1.5em} 
    
    \includegraphics[width=0.85\linewidth]{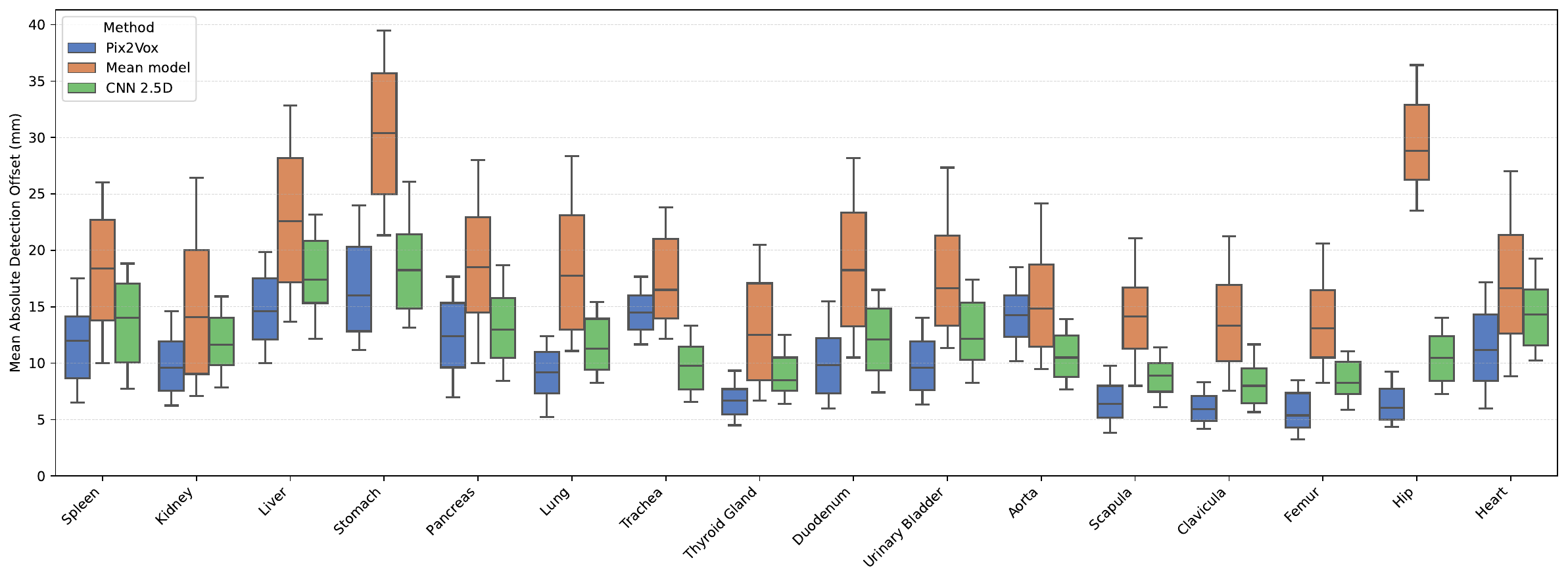}
    \caption{Detection offset for internal organs.}
    \label{fig_doe_others}
    
    \vspace{1.5em}

    \includegraphics[width=0.85\linewidth]{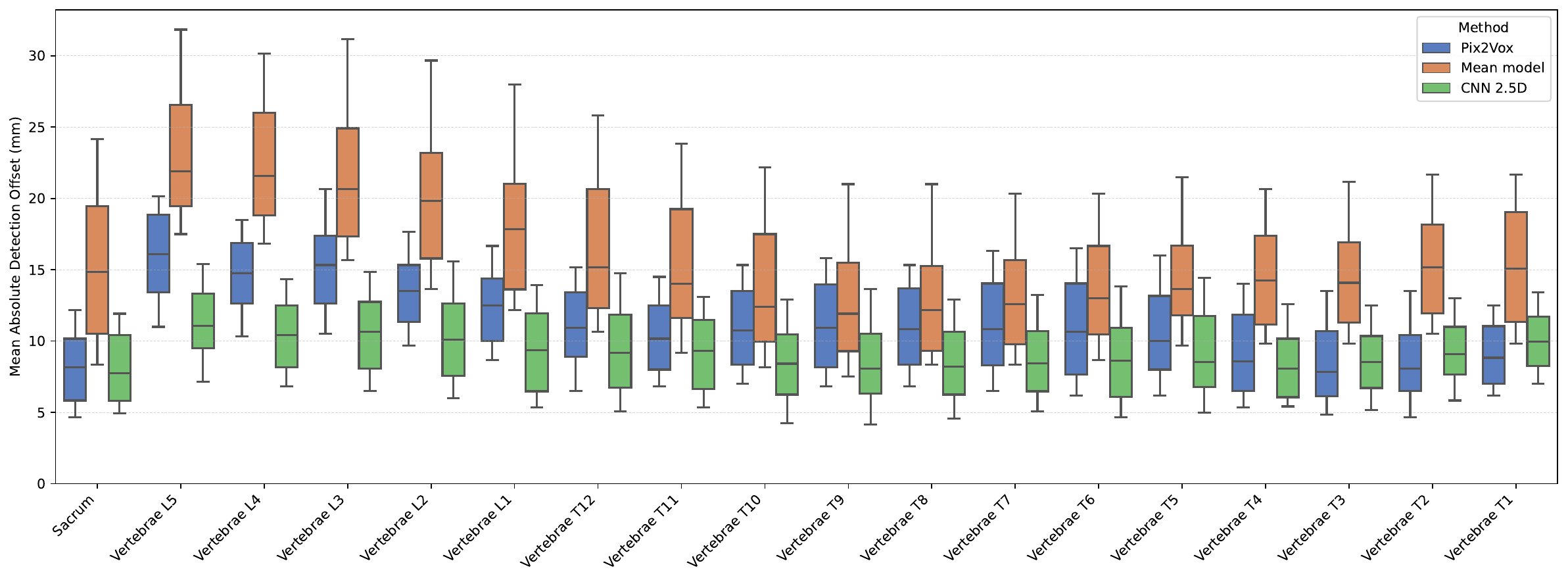}
    \caption{Detection offset for internal organs.}
    \label{fig_doe_vertebrae}
\end{figure*}

A detailed breakdown of performance metrics for each organ. Figures~\ref{fig_dice_others}--\ref{fig_doe_vertebrae} present the evaluation across all 41 anatomical structures.

\end{document}